\documentclass[sigconf,nonacm]{aamas} 
\usepackage{hyperref}   % Load it manually and early
\usepackage{hyperxmp}   % Safe now

\usepackage{balance} % for balancing columns on the final page
\usepackage{mathtools}
\setcopyright{none}
\acmConference[]{}
{}{}{}
\copyrightyear{}
\acmYear{}
\acmDOI{}
\acmPrice{}
\acmISBN{}
\usepackage{tikz}
\usetikzlibrary{calc,matrix}

\usepackage{bm}

\usepackage{graphics}
\usepackage{color}
\definecolor{rust}{rgb}{0.72, 0.25, 0.05}
\definecolor{princetonorange}{rgb}{1.0, 0.56, 0.0}

\definecolor{ann_color}{HTML}{42e6f5}

\title[Inclusive Fitness as a Key Step Towards More Advanced Social Behaviors in Multi-Agent Reinforcement Learning Settings]{Inclusive Fitness as a Key Step Towards More Advanced Social Behaviors in Multi-Agent Reinforcement Learning Settings}

\author{Andries Rosseau}
\affiliation{
  \institution{Vrije Universiteit Brussel}
  \city{Brussels}
  \country{Belgium}}
\email{andries.rosseau@vub.be}

\author{Raphaël Avalos}
\affiliation{
  \institution{Vrije Universiteit Brussel}
  \city{Brussels}
  \country{Belgium}}
\email{raphael.avalos@vub.be}

\author{Ann Nowé}
\affiliation{
  \institution{Vrije Universiteit Brussel}
  \city{Brussels}
  \country{Belgium}}
\email{ann.nowe@vub.be}

\begin{abstract}

The competitive and cooperative forces of natural selection have driven the evolution of intelligence for millions of years, culminating in nature's vast biodiversity and the complexity of human minds. Inspired by this process, we propose a novel multi-agent reinforcement learning framework where each agent is assigned a genotype and where reward functions are modelled after the concept of inclusive fitness. An agent’s genetic material may be shared with other agents, and our inclusive reward function naturally accounts for this. We study the resulting social dynamics in two types of network games with prisoner’s dilemmas and find that our results align with well-established principles from biology, such as Hamilton's rule. Furthermore, we outline how this framework can extend to more open-ended environments with spatial and temporal structure, finite resources, and evolving populations. We hypothesize the emergence of an arms race of strategies, where each new strategy is a gradual improvement over earlier adaptations of other agents, effectively producing a multi-agent \textit{autocurriculum} analogous to biological evolution. In contrast to the binary team-based structures prevalent in earlier research, our gene-based reward structure introduces a spectrum of cooperation ranging from full adversity to full cooperativeness based on genetic similarity, enabling unique non team-based social dynamics. For example, one agent having a mutual cooperative relationship with two other agents, while the two other agents behave adversarially towards each other. We argue that incorporating inclusive fitness in agents provides a foundation for the emergence of more strategically advanced and socially intelligent agents.

\end{abstract}


\newcommand{\BibTeX}{\rm B\kern-.05em{\sc i\kern-.025em b}\kern-.08em\TeX}

\begin{document}

%%% The following commands remove the headers in your paper. For final 
%%% papers, these will be inserted during the pagination process.

\pagestyle{fancy}
\fancyhead{}

%%% The next command prints the information defined in the preamble.

\maketitle 

%%%%%%%%%%%%%%%%%%%%%%%%%%%%%%%%%%%%%%%%%%%%%%%%%%%%%%%%%%%%%%%%%%%%%%%%

\section{Introduction}

Creating intelligent agents with the ability to adapt to a diverse set of challenges and environments is a prominent goal of artificial intelligence research. In the past decades, the field of single-agent Reinforcement Learning (RL)~\cite{sutton2018} has made great progress in developing agents capable of completing tasks provided in the form of a reward signal~\cite{mnih2015human, schulman2017proximal, haarnoja2018soft, mirowski2016learning, akkaya2019solving}. However, in traditional single-agent RL, once an agent has learned to master its given task, learning stops, since there is no further incentive for improvement. This makes it difficult to create agents which can solve a wide variety of complex tasks; an important characteristic of general intelligence. Creating extra handcrafted tasks for an agent to train and generalize on (i.e., transfer learning~\cite{taylor2009transfer}) can mitigate this, but this approach is resource intensive and still caps the final complexity that the agent can achieve. Recent work on procedurally generated tasks and environments~\cite{cobbe2020leveraging, wang2019paired, suarez2021neural} attempts to solve this problem, but creating adequate reward signals in the environments still remains costly.

Another problem is that an objective can be too complex to learn from scratch. In this case, the space of possible policies is too large to cover effectively with regular exploration strategies. This leads to an agent that cannot close the gap between its initial random behavior and solving the task, getting stuck in low-performing suboptima. Intermediate rewards (such as reaching checkpoints in a maze) can be created to help the agent learn the final overarching goal, but over-engineering the reward signal can lead to problems like specification gaming~\cite{krakovna2020specification, amodei2016concrete} and potentially limits the range and originality of the learned strategies. 

A slightly different approach is the use of curriculum learning~\cite{bengio2009curriculum, czarnecki2018mix, narvekar2020curriculum}, in which an agent gradually progresses from an easy environment (e.g., a small maze) towards more difficult ones (the full maze), similar to how human education works. Curriculum learning can provide an efficient way of learning complex tasks, but the final complexity is limited by the most advanced task, and creating suitable progressions in the curriculum is resource intensive. 

Improved exploration strategies can help an agent to avoid getting stuck in suboptimal states. Work in the field of intrinsic rewards proposes ways of overcoming a sparse reward signal. For example, curiosity driven RL~\cite{pathak2017curiosity, burda2018large, burda2018exploration} provides a self-supervised reward signal which promotes the exploration of previously unknown environment dynamics, often leading to the development of useful skills solely by following the intrinsic reward.

Nature, however, is not a single-agent system, but a multi-agent world full of evolving organisms. The competitive and cooperative forces of natural selection have driven the evolution of intelligence for many millions of years, culminating in nature's great biodiversity and the richness of our human minds. In nature, when a strategy with an increased fitness emerges, it changes the environment dynamics for others, creating a new set of challenges to adapt to.
% ~\cite{williams2018adaptation}
The agents that successfully adapt to these challenges have in turn improved their strategies, thereby again providing new challenges, and so forth. Less successful agents that are unable to keep up go extinct. Agents are therefore always at a similar level, and provide just the right amount of challenge for growth.
This can be applied to reinforcement learning as well: learning agents continuously improve, thereby pushing the others to adapt, leading to the emergence of a multi-agent \textit{autocurriculum}~\cite{leibo2019autocurricula}. Multi-agent autocurricula provide a scalable way for agents to explore a large strategy space by simply following the gradients of their experience, called ``exploration by exploitation''~\cite{Baker2020Emergent, leibo2019autocurricula}. 

In theory, an autocurriculum enables the possibility of unbounded growth for innovation, limited only by the strategy space of the environment and the agents' learning capacities. Autocurricula have formed the backbone for some of the most advanced forms of artificial intelligence known to date. In the form of self-play, it has led to agents with superhuman capabilities in the two-player zero-sum games of Backgammon~\cite{tesauro1995temporal}, Go, Chess and Shogi~\cite{silver2018general}, and the continuous real-time strategy game of StarCraft II~\cite{vinyals2019grandmaster}. In team-based competitive environments, it has led to agents beating the world champions in the real-time strategy game of Dota2~\cite{berner2019dota}, to human-level performance in a first-person 3D multiplayer game of capture-the-flag~\cite{jaderberg2019human}, and to an arms race in a 3D hide-and-seek game~\cite{Baker2020Emergent}, where several distinct strategic phases emerged, each requiring increasingly sophisticated forms of cooperation and tool use.

Our work fits in the tradition of the aforementioned works on multi-agent autocurriculum learning, aiming to create high levels of complexity starting from elegant, simple rules. However, the dynamics in previous work on multi-agent autocurricula are limited to either all competition, or in the case of predefined team setups, all cooperation and additional competition between teams. Humans -- like many other organisms in nature~-- are not that binary, instead showing a range of cooperative behavior. As the main contribution of this work, we propose the first steps towards a novel multi-agent autocurriculum inspired by biological evolution, where we construct an evolutionary aligned reward based on the fitness of an agent's genes. Since other agents potentially carry (pieces of) the same genetic material, the reward function incorporates the presence of these genes in other agents as well, by adding the other agents' individual payoffs weighted by a measure of their genetic relatedness. Our reward structure therefore leads to a spectrum of cooperation, based on relatedness. This spectrum can shift over time: for example, when the total population size grows, resources become scarce, incentivizing to favor closer relatives over agents carrying less of your genetic material. At any given time, the actual level of cooperation between agents is therefore determined by the size and content of the genotype population, and the actions of the present genotypes influence the gene pool that will occur in the future. This adds a novel social dimension to the multi-agent autocurriculum, which continuously challenges the agents to find new strategies that balance cooperation and defection appropriately. We argue that this multi-agent autocurriculum can lead to a continuous growth in strategic complexity, only bounded by the strategy space of the environment and the agents' learning capabilities.

The work most closely related to ours is that of Abrantes et al.~\cite{abrantes2020mimicking}, who are likewise inspired by evolutionary principles. Their reward function corresponds to the last of the three reward functions we propose for future work (Eq.~\ref{neural_combined_reward}). In their study, the authors demonstrate that agents can successfully learn to reproduce and survive in a team-based environment with dynamic population sizes. Although they also explore non-binary interaction settings, we believe their results do not yet provide a clear existence proof for the emergence of novel, non–team-based dynamics. While this is not the main focus of the present work, we hope that our future work section will further contribute to this discussion.

%%%%%%%%%%%%%%%%%%%%%%%%%%%%%%%%%%%%%%%%%%%%%%%%%%%%%%%%%%%%%%%%%%%%%%%%

\section{Methods}\label{methods}

\subsection{Information stability}\label{information_stability}
We propose a general definition of fitness as the stability of an \textit{information state} with regards to its environment. The more stable an information state is, the longer it will keep existing. For example, galaxies or diamonds are information states with a high fitness in the realm of physics. When considering complex organic molecules on a primeval Earth, the strategy of producing a copy of oneself before being destroyed (i.e., replicating) turned out to be a particularly stable one, drastically lengthening the existence of one's information state. Yet, every so often, an error occurs in the copying process, known as a mutation. When a mutation is beneficial, it leads to an improvement in fitness, which is favored by natural selection. At the same time, however, a mutation also changes the information state. Mutations, coupled with natural selection, generally lead to a gradual shift in the population of replicators, towards increasingly stable (fit) information states. Replicators, which exist today in the form of DNA, have built an astonishing set of ingenious organisms around themselves to help them survive and replicate. 

In biology, the parts of DNA that code for the observable traits (i.e., the phenotype) of an organism are the genes\footnote{One could argue this is an oversimplification, for instance because it ignores regulatory elements and epigenetic effects.}, and together they constitute the genotype. The genotype represents the complete information state upon which selection ultimately acts\footnote{Strictly speaking, selection acts on phenotypic variation, but because only the genotype is inherited, evolutionary change is reflected in changes in genotypes mediated by their phenotypic effects.}. To translate our definition of fitness into an evolutionary aligned reward function for reinforcement learning, we implement an abstract version of genetics in our agents. We assign an agent $i$ with an abstract genotype $\bm{g}_i$, which is a sequence of $n$ genes where each gene locus/index $k\!\in\![1,n]$ contains a gene $g_i^k$. Different integer values for $g_i^k$ then represent different gene variants, where in principle every gene locus can have an undetermined amount of gene variants. 

We propose a metric of information similarity between genotypes to quantify their relatedness. In information theory, the Hamming distance $H(\bm{s_1}, \bm{s_2})$~\cite{hamming1950error} between two sequences $\bm{s_1}$ and $\bm{s_2}$ is the number of positions at which corresponding entries are different, measuring the amount of substitutions (`bit flips') needed to change one sequence back into the other. Starting from the normalized Hamming distance, we derive a similarity metric, expressing the genetic relatedness between two agents as a real number between 0~and~1, which we name the \textit{Hamming similarity}. Considering two agents $i$ and $j$, the Hamming similarity is defined as:
\begin{align}
     h(\bm{g}_i, \bm{g}_j)\, \equiv \, 1 - \frac{1}{n}\, H(\bm{g}_i, \bm{g}_j)\, =\, \frac{1}{n} \sum_{k=1}^n \delta(g_i^k, g_j^k) \,,
     \label{hamming}
\end{align}
\noindent where $\delta(\cdot\,,\cdot)$ is the Kronecker delta. Note that this metric is defined when both genotypes have the same length. In the case of different genotype lengths, we could use the Damerau-Levenshtein distance~\cite{brill2000improved}, an extension of the Hamming distance which takes into account information deletions and insertions as well.

\subsection{Inclusive fitness}\label{ev_al_rew}

\subsubsection{Inclusive reward function}
Since an agent's genetic material can be present in others as well, helping agents which are genetically related should also be promoted by our reward function. We therefore modify the reward of each agent $i$ by adding the rewards of the other agents as well, multiplied by their Hamming similarity $h$ (Eq.~\ref{hamming}). We call this modified reward the \textit{inclusive reward}, after the concept of inclusive fitness~\cite{rogers2021inclusive, hamilton1964}, which posits that under the right circumstances, natural selection favors organisms that help their genetic relatives. We define the inclusive reward $r^{*}$ as:
\begin{align}
\label{eq:inclusive_reward}
    r^{*}_i \equiv \sum_{j} h(\bm{g}_i, \bm{g}_j) r_j = r_i + \sum_{j \neq i}  h(\bm{g}_i, \bm{g}_j) r_j 
\end{align}

\begin{figure}[]
    \centering
    \includegraphics[width=8.3cm]{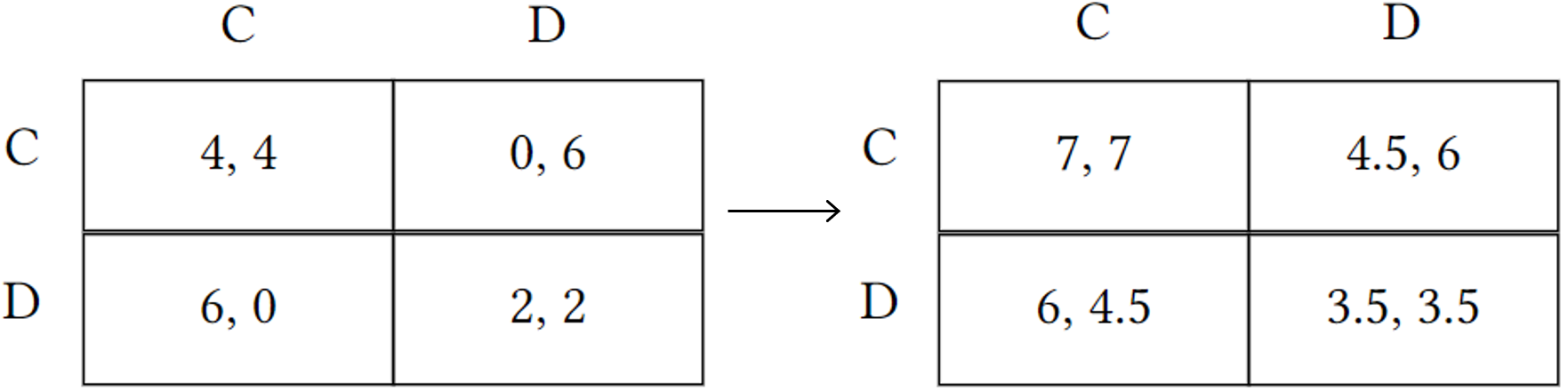}
    \caption{A prisoner's dilemma played by two players with genotypes [1,~1,~1,~1] and [1,~1,~1,~0] becomes a harmony game under the inclusive reward.}
    \label{fig:pd_harmony}
\end{figure}

An illustrative example of an inclusive reward is given in Figure~\ref{fig:pd_harmony}, where we consider two agents playing a prisoner's dilemma~\cite{kuhn2008prisoner}. Both agents can either cooperate (C), or defect (D). A rational agent will choose to defect, since that action always provides more payoff, regardless of the action of its opponent. However, when both agents defect, they are worse off than had they both cooperated, leading to the dilemma. The dynamics change drastically when we introduce genes and our inclusive reward function. If we consider that the genotypes of the row player and the column player are [1,~1,~1,~1] and [1,~1,~1,~0], respectively, they have a Hamming similarity of $\frac{3}{4}$. The immediate payoff of an agent, which we will call the \textit{individual payoff} $P$, indicates an agent's individual fitness, regardless of others. But an agent's action influences the payoff of its opponent, which carries three of its genes. The total \textit{inclusive} reward of the row player then becomes $P_{\text{row}} + \frac{3}{4} P_{\text{column}}$, with a symmetric expression for the column player. Therefore, from the perspective of the genotypes, the prisoner's dilemma of Fig.~\ref{fig:pd_harmony} effectively becomes a harmony game~\cite{zizzo2002measurement}, where the only Nash Equilibrium is to both cooperate. 

\subsubsection{General prisoner's dilemma}\label{generalpris}
In a general prisoner's dilemma, $b$ is the benefit provided to the other by cooperating, and $c$ is the cost for cooperation. The payoff matrix is given by:
\vspace{-3mm}
\begin{table}[h!]
\begin{tabular}{lcc}
                       & C                                 & D                              \\ \cline{2-3} 
\multicolumn{1}{l|}{C} & \multicolumn{1}{c|}{$b-c$, $b-c$} & \multicolumn{1}{c|}{$-c$, $b$} \\ \cline{2-3} 
\multicolumn{1}{l|}{D} & \multicolumn{1}{c|}{$b$, $-c$}    & \multicolumn{1}{c|}{0, 0}      \\ \cline{2-3} 
\end{tabular}
\end{table}
\vspace{-2mm}

\noindent From this general payoff matrix, we can derive two inequalities that need to be satisfied for cooperation to be favored under the inclusive reward: $c<hb$ and $hc<b$ (with $h$ the Hamming similarity between the two players). We do not consider the second inequality, which is simply a consequence of the first, since $h\! \in \! [0, 1]$. This first inequality turns out to be equivalent to Hamilton's rule~\cite{hamilton1964} from biology, which posits that an cooperative trait can persist if the benefit $b$, multiplied by the relatedness $r$, exceeds the cost $c$.

\section{Cooperation on networks}

Our evolutionary aligned reward function should incentivize an agent to maximize the fitness of its genetic material, which can be present in other agents as well. Therefore, we defined an inclusive reward (Eq.~\ref{eq:inclusive_reward}) which adds the individual rewards by weighing them with the Hamming similarity defined in Eq.~\ref{hamming}. In this section, we study the properties of this inclusive reward by focusing on two settings where independent Q-learners~\cite{watkins1992q} play two-player prisoner's dilemmas on networks. Self-interested agents often fail to cooperate in prisoner's dilemmas due to the dominance of the defective strategy over cooperation. In nature, however, many organisms have evolved stable cooperative strategies~\cite{hamilton1964}. The goal of these experiments is to show the emergence and stability of cooperation in environments where agents try to maximize the fitness of their genetic material.

\begin{figure}[]
    \centering
    \includegraphics[width=6cm]{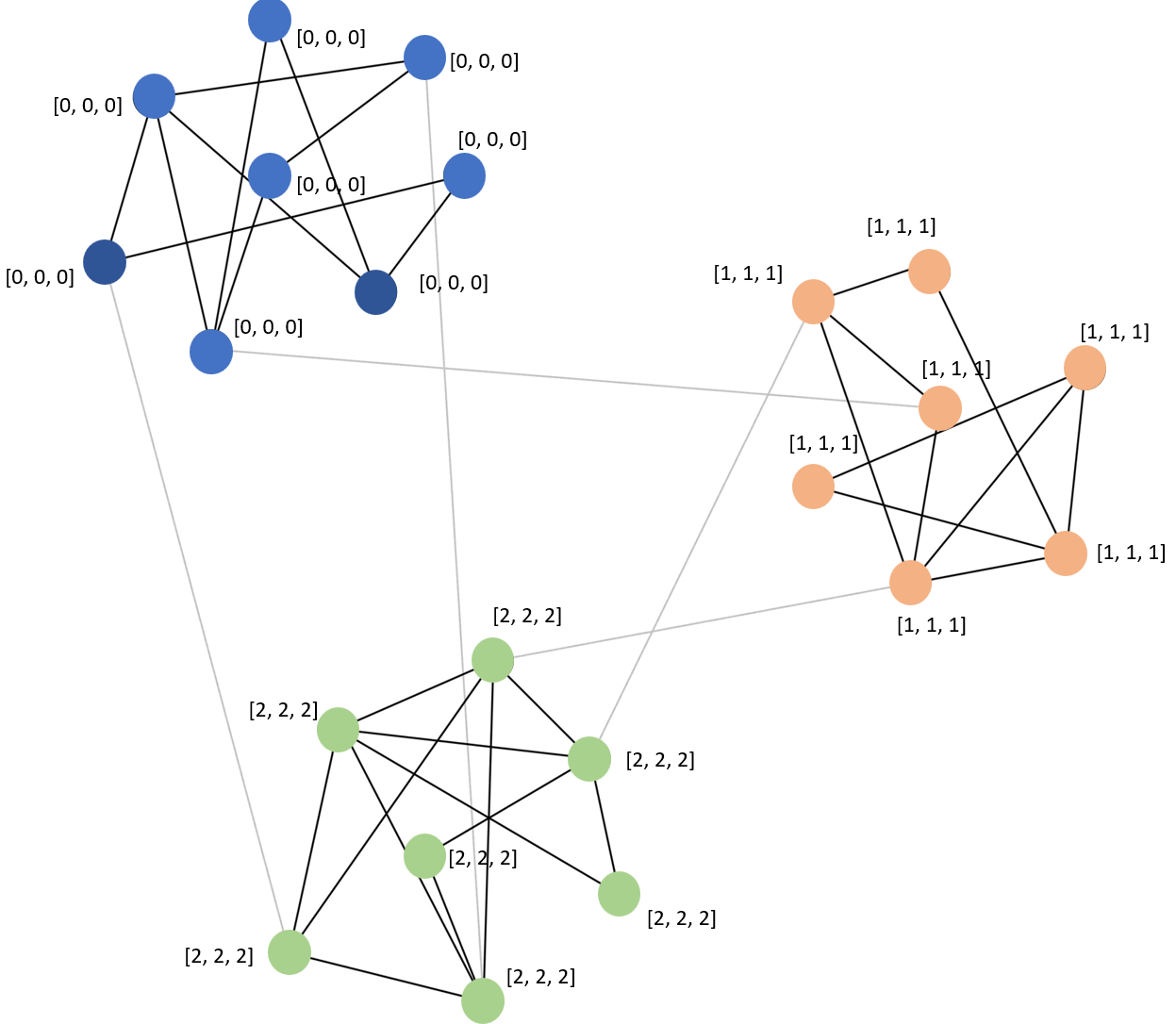}
    \caption{Example network with community structure. Here, the probability of a connection between agents inside a community is $p_{in} = 0.9$, while the probability for agents between communities is $p_{out} =0.1$. All three communities represent a separate genotype. Figure adapted from~\cite{girvan2002community}.}
    \label{fig:community}
\end{figure}

\subsection{Experiments}

% \subsubsection{Opponent discrimination}
\subsubsection{Opponent discrimination}
A first experiment considers fully connected networks where agents can recognize each other, often referred to as \textit{opponent discrimination}. This means that an agent knows which opponent it is playing, but it does not know what genotype the opponent has, nor does it remember anything of what the agent did in the past; it only bases its action on a learned behavior for that opponent (i.e., a Q-table). The setup is based on the evolution of sensing organs, which provide an organism the ability to observe the \textit{phenotype} of other organisms in the environment, but not directly its genotype. Senses such as vision are crucial for animals, and over many generations led to intuitive recognition of offspring, or the avoidance of predators~\cite{williams2018adaptation}, examples which we intend to capture with our setup. Opponent discrimination is implemented in our agents as a Q-table where every state corresponds to a different opponent on the network, and the agents receive each time step as an observation which opponent they are playing.

\subsubsection{Limited dispersal}
Our second experiment gives agents no opponent discrimination, which means agents have only one strategy for all interactions. Instead, we look at the effect of limited dispersal (often also called population viscosity~\cite{hamilton1964, hamilton1972altruism}) on the emergence of cooperation between independent Q-learners under our inclusive reward. Under the limited dispersal hypothesis, it is assumed that organisms do not disperse far from their birth place, making them more likely to interact with genetic relatives. 

To model limited dispersal, we move from fully connected networks to random partition networks~\cite{fortunato2010community} which have community structure~\cite{girvan2002community}. Random partition networks are constructed starting from predefined groups of nodes that form (still unconnected) communities. Nodes that belong to the same community are connected with probability $p_{in}$, and nodes between communities with $p_{out}$. We define a \textit{dispersal coefficient} $\eta \equiv p_{out}/p_{in} \in [0,1]$, denoting the strength of the network dispersal. Every node in a community has the same genotype. This means that agents with similar genotypes are more likely to be connected than others (Fig.~\ref{fig:community}). The influence of network structure on the emergence of cooperation has been well-studied in evolutionary game theory~\cite{ohtsuki2006simple, santos2005scale, nowak1992evolutionary, szabo2007evolutionary}. Here, we provide an alternative approach of modelling the strategies with reinforcement learning, where we study the resulting dynamics under an evolutionary aligned (inclusive) reward.

\subsubsection{Reward}
The payoff matrix of the prisoner's dilemma provides the individual fitnesses that each agent receives under their combined actions. We again use these individual fitnesses to construct our inclusive reward, according to Eq.~\ref{eq:inclusive_reward}. After every interaction, a player $i$ uses its individual payoff $P_i$ and the opponent's payoff $P_j$ to determine its inclusive reward $r^{*}_i$:
\begin{align}
    r^{*}_{i} = P_i + h(\bm{g}_i, \bm{g}_j) P_j \,.
\end{align}

In both the opponent recognition and the limited dispersal experiment, our Q-learners try to optimize their myopic inclusive reward (meaning a bandit-like discount factor of $0$), similar to how generational fitness is often defined in evolutionary game theory~\cite{szabo2007evolutionary, ohtsuki2010evolutionary, ohtsuki2006simple, fudenberg2006evolutionary}. Players pick and update their Q-values according to an $\epsilon$-greedy scheme with exponentially decaying exploration. 

\subsection{Results}

\begin{figure}[]
    \centering
    \includegraphics[width=8.4cm]{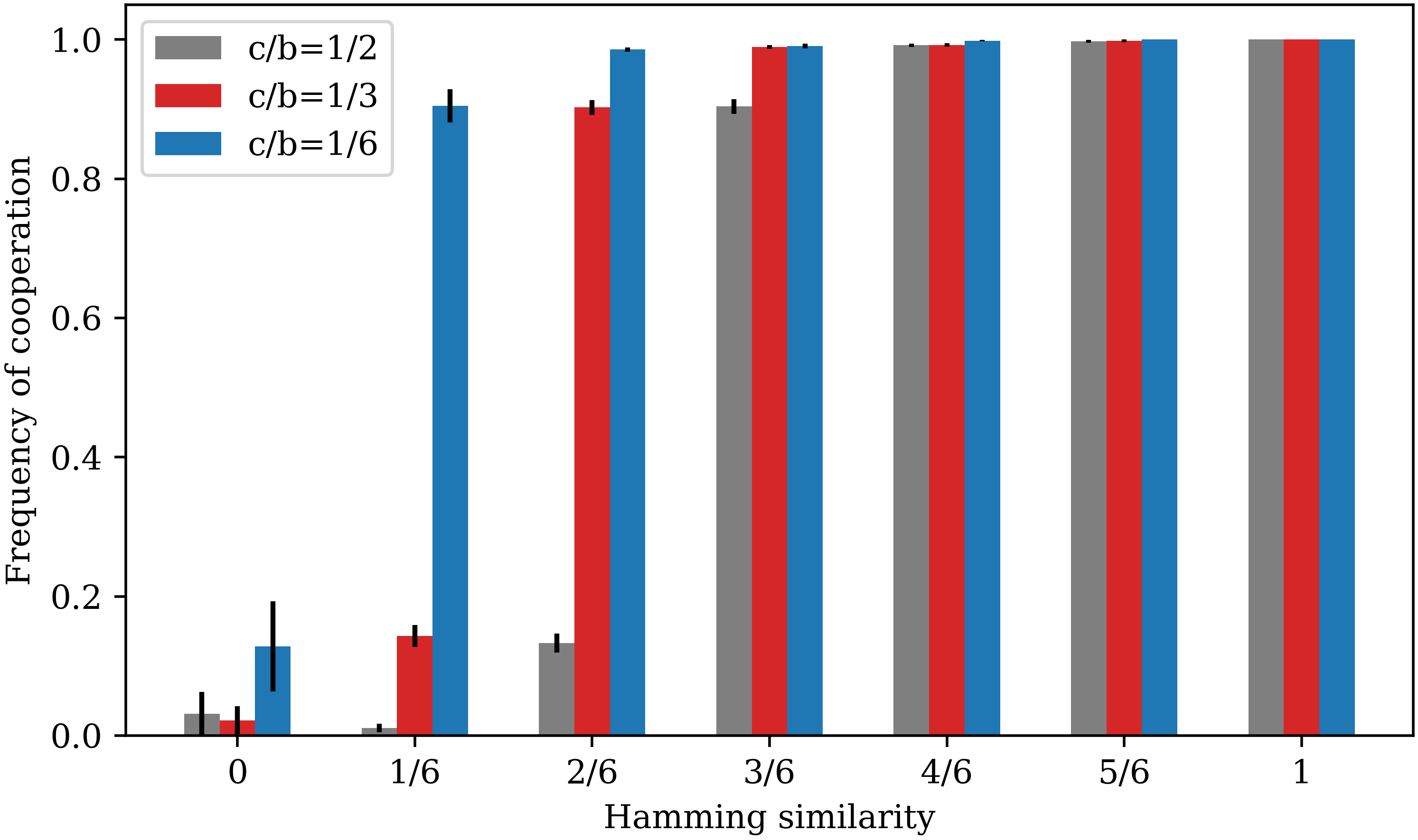}
    \caption{Frequency of cooperation in function of the Hamming similarity with the opponent on a fully connected network (genotype length 6, 2 variants per gene) with 64 agents, one per unique genotype. The cost-benefit payoff ratio $c/b$ influences the Hamming similarity threshold at which agents start cooperating, matching Hamilton's rule~\cite{hamilton1964}.}
    \label{fig:recognition}
\end{figure}

\begin{figure}[]
    \centering
    \includegraphics[width=8.4cm]{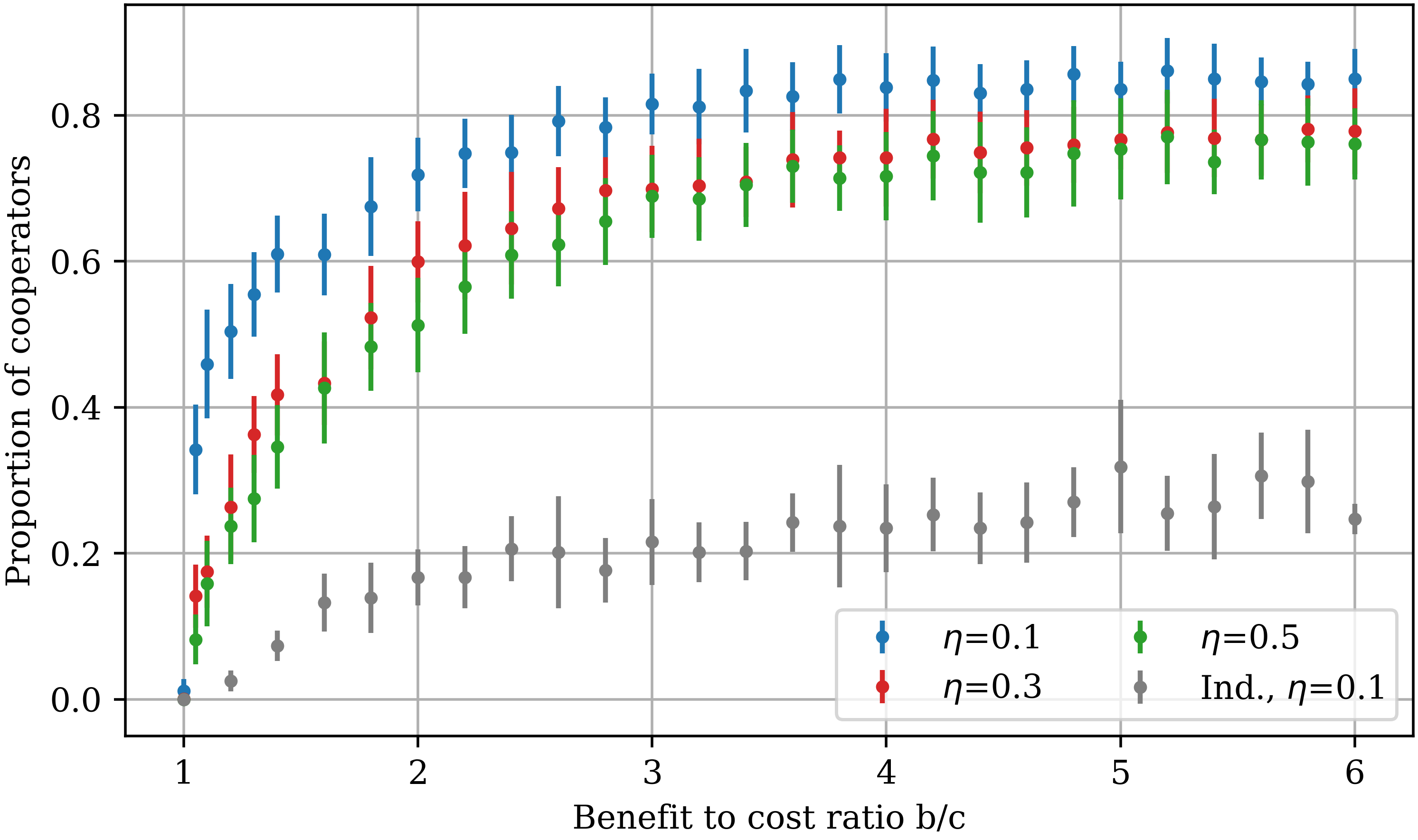}
    \caption{Proportion of cooperators in the (converged) population in function of the benefit to cost ratio $b/c$ in a random partition network (genotype length 3, 2 variants per gene). Each community represents one of $8$ unique genotypes, with 8 agents per genotype/community. $\eta$ is the network dispersal coefficient. $\langle k \rangle \! = \! 9$ for all experiments, which together with $\eta$ determines $p_{in}$ and $p_{out}$ (Eq.~\ref{eq:avk}). Blue, red and green values show the proportion of cooperators under the inclusive reward for different dispersal coefficients. Gray values show results without inclusiveness. Cooperation increases under the inclusive reward and under small dispersal coefficients.}
    \label{fig:dispersal}
\end{figure}

\subsubsection{Opponent discrimination}
We use a fully connected network for opponent discrimination, where each node represents an agent. We consider genotypes of length 6, with 2 gene variants per gene locus. We create one agent for every possible genotype, thereby making the network symmetric for all agents, for a total of $2^6=64$ combinations. All Q-tables are initialized to zero. Initial populations of all defectors, all cooperators, and mixtures were tried as well, but did not influence the results. Each time step, agents play one prisoner's dilemma against all of their opponents simultaneously, including itself, where the individual payoffs are defined by the benefit $b$ and the cost $c$ ($c$ is fixed at 1, while we vary $b$). Figure~\ref{fig:recognition} shows the resulting frequencies (averaged over the agents) of agents cooperating with their opponents, visualized with respect to the Hamming similarity for three values of $c/b$. The results match with Hamilton's rule~\cite{hamilton1964}, which as noted in section~\ref{generalpris} predicts the spread of a cooperative strategy if $c/b < h$. Results without inclusive rewards (not shown in Figure~\ref{fig:recognition}) led to all defection.

\subsubsection{Limited dispersal}
We consider a random partition network, where agents have genotypes of length 3, with 2 variants per gene. A community of 8 nodes is created per unique possible genotype, i.e. 8 communities of 8 nodes. In figure~\ref{fig:dispersal}, we consider three values of the dispersal coefficient $\eta$, and measure the proportion of cooperators that emerge in the network after convergence, with respect to the benefit to cost ratio $b/c$. Although we vary $\eta$, we keep the average degree fixed at $\langle k \rangle = 9$ to avoid that a variation in $\eta$ leads to a variation in average degree, since it is known that varying the average network degree can drastically influence the spread of cooperation~\cite{ohtsuki2006simple, santos2005scale, nowak1992evolutionary}. We keep $\langle k \rangle$ steady by deriving $p_{in}$ and $p_{out}$ through the following relation for $\langle k \rangle$, $\eta$ and $p_{in}$:
\begin{align}
    \langle k \rangle = 9 = 7 p_{in} + 56 p_{out} = (7 + 56 \, \eta) \, p_{in}.
    \label{eq:avk}
\end{align}

The results from Figure~\ref{fig:dispersal} with inclusive reward show higher proportions of cooperation than with individual rewards, even though some level of cooperation can emerge without inclusiveness. Small dispersal coefficients and large benefit-to-cost ratios also lead to higher levels of cooperation, which matches with our prediction based on limited dispersal theory.

\begin{figure*}[]
    \centering
    % \raggedright
    \includegraphics[width=14.cm]{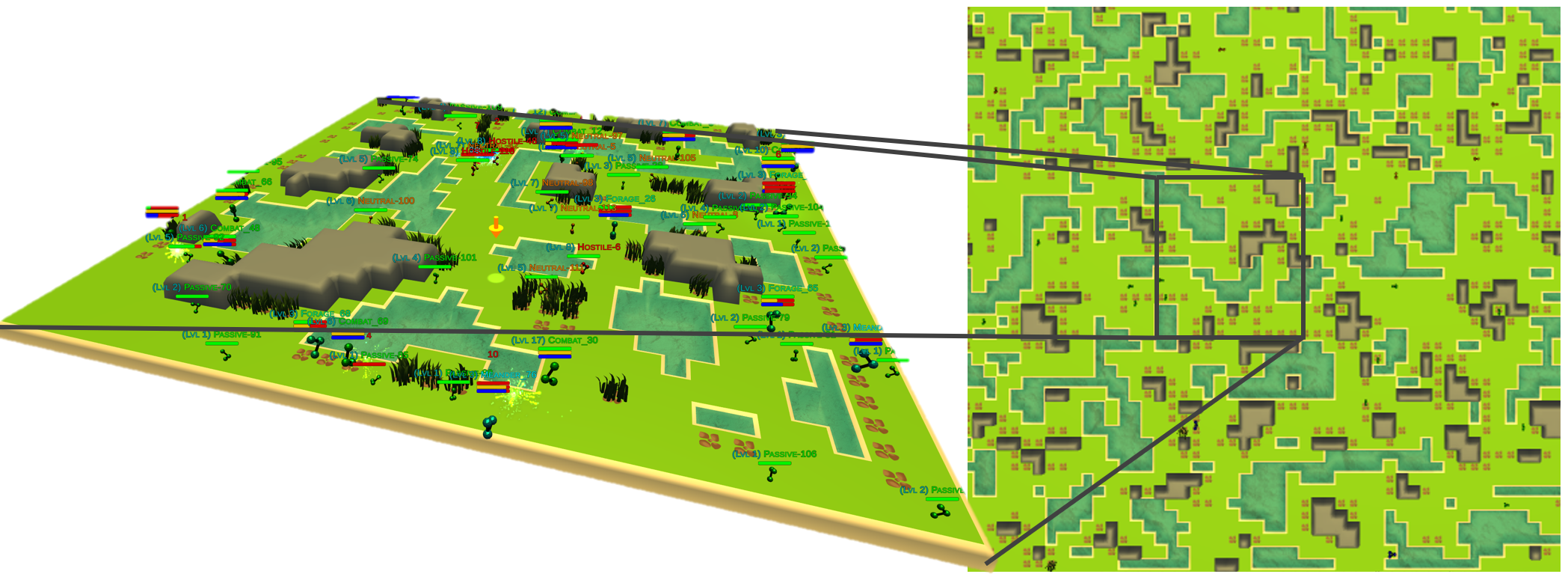}
    \caption{Impression of the Neural MMO environment~\cite{suarez2021neural}. Rocks, water, grass and forest cover the area. Agents, represented by three connected nodes, gather food and water, and engage in combat with each other or with environmental threats.}
    \label{fig:neuralmmo}
\end{figure*}

% \vspace{-2mm}

\section{Markov games}

In future work, we intend to move beyond the limitations of networks and abstract matrix games, into temporally and spatially extended Markov games~\cite{littman1994markov}. Our main focus will be on the Neural MMO environment~\cite{suarez2021neural} (Fig.~\ref{fig:neuralmmo}). Neural MMO is a multi-agent video game environment in which agents survive by gathering resources such as water and food in a rugged environment. The agents can also engage in combat against other agents or with threats from the environment itself. Neural MMO is open-source, and serves as a customizable platform for multi-agent intelligence research, where agent configurations, reward functions, environment layout, resources, etc. can be rewritten and adjusted to suit the purposes of our research.

The goal of the previous network experiments was to examine the evolution of cooperation under a genetic inclusive reward function. Now, by moving to more open-ended environments like Neural MMO, we will test our hypothesis that an inclusive reward can induce an autocurriculum that can lead to the emergence of increasingly complex and socially intelligent strategies. In contrast to our network experiments, agents in the Neural MMO environment can move around, which leads to a non-stationary or dynamic `network' of interactions. Moreover, where matrix games previously provided strategies as directly accessible atomic actions, agents now need to learn to implement high-level strategies with policies that execute a sequence of many low-level actions. Rewards cannot be matched with one clear causal action anymore, resulting in the credit-assignment problem~\cite{sutton1984temporal}, and the concept of a general well-defined interaction between two agents ceases to exist, since an agent's actions can now influence events removed in distance as well as time. Moreover, the most rewarding and innovative strategies are not directly accessible anymore, but have to be discovered and learned. Strategies of cooperation and defection can take many forms and decisions do not always need to occur at the same moment: some information about what a player is starting to do can change another player’s actions in the process.

\subsection{Evolutionary aligned rewards}\label{rewardfunction}
In the previous games, fitness was simply represented by the payoffs in the payoff matrix, and then used to create the inclusive reward. Here, computing the fitness of an agent is not as straightforward anymore, and we need to propose a mathematical expression of fitness to create our evolutionary aligned reward, in line with our definition of fitness as a measure of stability or longevity of an information state (section~\ref{information_stability}). 

\paragraph{Longevity reward}
To optimize for longevity, an agent receives a reward of $+1$ for every time step that at least one copy of its genotype is still alive. Since other agents can share its genes, the agent receives an additional inclusive reward of $+1$ for every related genotype alive, multiplied by the Hamming similarity. This `longevity' reward $r^{L}$ for agent $i$ at time step $t$ then becomes
\begin{align}
    r^{L}_{i, t} \coloneqq \sum_{\bm{g}_j \in G_t} h(\bm{g}_i, \bm{g}_j),
    \label{neural_reward}
\end{align}
where $G_t$ is the set of \textit{unique} genotypes alive at $t$.

This reward optimizes for the long term survival of an agent's genes, while leaving the strategy of how to accomplish this long-term survival open for the agents to discover. 

\paragraph{Replication reward}
In the field of biology, fitness is often defined as the expected number of offspring an organism produces~\cite{mills1979propensity, lerner1958genetic, waddington1968towards}. Therefore, we also propose a reward function which directly promotes the maximization of the \textit{amount} of shared genetic material that is present in the environment. Now, an agent $i$ receives a reward of +1 for every newborn, and -1 for every agent that dies, which we again weigh by the Hamming similarity to include the potential presence of an agent's genes in other agents as well. This `replication' reward $r^{R}$ for agent $i$ at time step $t$ is then given by
\begin{align}
    r^{R}_{i, t} = \sum_{j \in J_t} h(\bm{g}_i, \bm{g}_j) - \sum_{j \in J_{t-1}} h(\bm{g}_i, \bm{g}_j),
    \label{neural_replication_reward}
\end{align}
where the first sum is over the group of agents $J_t$, alive at time step $t$, and the second sum over the agents from the previous time step $t\!-\!1$, where every agent is weighted by its Hamming similarity.

\paragraph{Combined reward}
Given that the longevity reward -- excluding the possibility of immortal organisms -- will eventually lead to the discovery of replication as a crucial part of an agent's strategy, we can extend our longevity reward with the concept behind the replication reward, providing a reward signal that combines the best of both. We define this `combined' reward $r^{C}$ for agent $i$ at time step $t$ as
\begin{align}
    r^{C}_{i, t} = \sum_{j \in J_t} h(\bm{g}_i, \bm{g}_j),
    \label{neural_combined_reward}
\end{align}
where we sum over the group of agents $J_t$ that are alive at time step $t$. The combined reward simply gives a positive reward (the Hamming similarity) when an agent which carries the same genetic material is alive for one more time step. The difference between the original longevity reward and the combined reward is subtle. In the former, we sum over the unique \textit{genotypes} alive, for which all agents that carry a specific genotype only count for one positive reward, since only one copy is required for the information state to be alive. In the latter case, however, we also take into account how many copies of those unique genotypes there are. Note that the replication reward is equal to the difference in the combined reward over two subsequent time steps.

When performing our experiments, we intend to try all three reward functions and study their properties.

\subsection{Rules of the game}
The rules of the game in our intended adaption of the Neural MMO environment are as follows. One health point is subtracted every time step from an agent's total health. To replenish health, an agent can consume water or food, which has to be gathered by standing near a pool or walking through a forest area. Besides foraging, agents can engage in combat, and attack each other when they are in striking distance for high damage, or with projectiles from afar, which deal less damage but are safer. The action space of an agent consists of movement actions, attack actions, and very importantly, an action that makes the agent reproduce. For the reproductive process, we propose to implement a simple system: when an agent decides to reproduce, it gives 1/4th of its health and resources to its offspring. More elaborate schemes are of course possible. Here, a strategy that reproduces fast will have more offspring, but produces agents that are generally weaker and more prone to getting killed early on. Agents will either learn one optimal reproductive strategy, or perhaps develop niches like some agents focusing on multiplication while others focus on strong individuals (e.g., predator-prey dynamics).

The resources in Neural MMO are not endless. When resources are plentiful, each reward function will promote cooperation between any two agents that share at least one gene, and in principle induce indifference for agents that share none. However, when the population size grows towards the carrying capacity of the environment, resources become scarce, and a tension arises between helping closer relatives and more distant ones. Helping agents with a lower genetic similarity would mean the consumption of resources that could be used to help agents with higher similarity instead, so we expect to see a non-stationary spectrum of cooperation emerge.

Moreover, novel, non team-based social dynamics can emerge. For instance, if agents of type A share 60\% of their genes with agents of types B and C (i.e., $h=0.6$), but agents of types B and C only share 20\% of their genes ($h=0.2$), it is possible that resource scarcity incentivizes cooperation between genotypes only when $h>0.5$. In this case, agents of type B and C are adversaries and will likely learn to attack each other, while agents of type A are in principle in a cooperative relation with both B and C. This could lead to novel behavior from type A agents, such as mediating between types B and C (given access to a proper mechanism for this, e.g., through gifting, or through being able to periodically restrict access to resources), or alternatively forming a coalition with one randomly chosen type and help extinguish the remaining agent type (even though it had no direct incentive to do so) by realizing it prefers its own type to be alive ($h=1$) plus one other complete type ($h=0.6$) over types B and C annihilating each other, which would lead to less inclusive reward since \textit{none} of the $h=0.6$ types are now alive.

\subsection{Experimental setup}
\subsubsection{Training}
To train our agents, we will move from tabular Q-learning to Deep Reinforcement Learning (Deep RL) with Proximal Policy Optimization (PPO)~\cite{schulman2017proximal} and Long Short-Term Memory (LSTM) layers~\cite{hochreiter1997long} to enable our agents to reach the necessary strategic complexity. To save computational resources, one could use a parameter-sharing scheme where all agents share the same neural network weights, but where the policy is conditioned on a unique genotype identifier, provided to the network as a dedicated part of the observation state. This is a common strategy in multi-agent RL~\cite{Baker2020Emergent, avalos2021local} which allows for specialization of strategies, where all the specializing strategies are condensed in one (large) neural network (i.e., a function approximation of different function approximators). The identifiers therefore allow the neural network to learn different policies for each agent. The alternative is to store and update separate policy networks for every agent. 

\subsubsection{Evolving the world}
We start the game with one agent, carrying a single genotype. Once this agent learns to reproduce, its offspring carries the same genotype. However, with a probability $\mu$, each of its genes can mutate to another gene variant. This creates a new species, which also carries a new unique policy identifier and can therefore grow into a new strategy. So far, genes only influence the behavior of agents through the reward function, but one could also explore genes that express properties of the agents themselves, such as in-game statistics like maximal health or combat strength. 

There are no predefined generations; each agent can reproduce at any time step, making the world and the population in it grow organically.

\section{Discussion}

Our experiments on networks with opponent discrimination and limited dispersal match well-established biological principles, such as Hamilton's rule and limited dispersal theory. The results hint at the the potential of our inclusive reward function for the emergence of dynamic social structures not limited to only full cooperation or competition. We believe that a more open-ended environment will allow us to present an empirical proof that our evolutionary aligned reward functions (section~\ref{rewardfunction}) can provide a continuous incentive for progress towards increasingly complex strategies within non-stationary social structures. We only expect to observe a \textit{direction} towards increasingly complex strategies; no specific strategic properties are hypothesized, except the maximization of genetic fitness. Still, the open-ended setting will likely lead to behavior that is interpretable in an evolutionary context. Moreover, our reward structure is not limited to settings like Neural MMO, but has the potential to be applied in a plethora of multi-agent reinforcement learning settings where non-stationary coalitions and expressive social dynamics play an important part.

In conclusion, we have outlined a translation of inclusive fitness into multi-agent reinforcement learning, thereby providing a viable approach for creating generally capable and socially intelligent agents, starting from simple rules. We propose that in sufficiently rich environments, our inclusive reward structure has the potential to lead to a high strategic complexity, where agents will learn to balance cooperative and competitive incentives well.

\balance{}
\bibliographystyle{ACM-Reference-Format} 
\bibliography{bibliography}


\begin{thebibliography}{50}

%%% the Web address.

\ifx \showCODEN    \undefined \def \showCODEN     #1{\unskip}     \fi
\ifx \showDOI      \undefined \def \showDOI       #1{#1}\fi
\ifx \showISBNx    \undefined \def \showISBNx     #1{\unskip}     \fi
\ifx \showISBNxiii \undefined \def \showISBNxiii  #1{\unskip}     \fi
\ifx \showISSN     \undefined \def \showISSN      #1{\unskip}     \fi
\ifx \showLCCN     \undefined \def \showLCCN      #1{\unskip}     \fi
\ifx \shownote     \undefined \def \shownote      #1{#1}          \fi
\ifx \showarticletitle \undefined \def \showarticletitle #1{#1}   \fi
\ifx \showURL      \undefined \def \showURL       {\relax}        \fi
% invisible to TeX
\providecommand\bibfield[2]{#2}
\providecommand\bibinfo[2]{#2}
\providecommand\natexlab[1]{#1}
\providecommand\showeprint[2][]{arXiv:#2}

\bibitem[\protect\citeauthoryear{Abrantes, Abrantes, and Oliehoek}{Abrantes et~al\mbox{.}}{2020}]%
        {abrantes2020mimicking}
\bibfield{author}{\bibinfo{person}{Jo{\~a}o~P Abrantes}, \bibinfo{person}{Arnaldo~J Abrantes}, {and} \bibinfo{person}{Frans~A Oliehoek}.} \bibinfo{year}{2020}\natexlab{}.
\newblock \showarticletitle{Mimicking evolution with reinforcement learning}.
\newblock \bibinfo{journal}{\emph{arXiv preprint arXiv:2004.00048}} (\bibinfo{year}{2020}).
\newblock


\bibitem[\protect\citeauthoryear{Akkaya, Andrychowicz, Chociej, Litwin, McGrew, Petron, Paino, Plappert, Powell, Ribas, et~al\mbox{.}}{Akkaya et~al\mbox{.}}{2019}]%
        {akkaya2019solving}
\bibfield{author}{\bibinfo{person}{Ilge Akkaya}, \bibinfo{person}{Marcin Andrychowicz}, \bibinfo{person}{Maciek Chociej}, \bibinfo{person}{Mateusz Litwin}, \bibinfo{person}{Bob McGrew}, \bibinfo{person}{Arthur Petron}, \bibinfo{person}{Alex Paino}, \bibinfo{person}{Matthias Plappert}, \bibinfo{person}{Glenn Powell}, \bibinfo{person}{Raphael Ribas}, {et~al\mbox{.}}} \bibinfo{year}{2019}\natexlab{}.
\newblock \showarticletitle{Solving rubik's cube with a robot hand}.
\newblock \bibinfo{journal}{\emph{arXiv preprint arXiv:1910.07113}} (\bibinfo{year}{2019}).
\newblock


\bibitem[\protect\citeauthoryear{Amodei, Olah, Steinhardt, Christiano, Schulman, and Man{\'e}}{Amodei et~al\mbox{.}}{2016}]%
        {amodei2016concrete}
\bibfield{author}{\bibinfo{person}{Dario Amodei}, \bibinfo{person}{Chris Olah}, \bibinfo{person}{Jacob Steinhardt}, \bibinfo{person}{Paul Christiano}, \bibinfo{person}{John Schulman}, {and} \bibinfo{person}{Dan Man{\'e}}.} \bibinfo{year}{2016}\natexlab{}.
\newblock \showarticletitle{Concrete problems in AI safety}.
\newblock \bibinfo{journal}{\emph{arXiv preprint arXiv:1606.06565}} (\bibinfo{year}{2016}).
\newblock


\bibitem[\protect\citeauthoryear{Avalos, Reymond, Now{\'e}, and Roijers}{Avalos et~al\mbox{.}}{2022}]%
        {avalos2021local}
\bibfield{author}{\bibinfo{person}{Rapha{\"e}l Avalos}, \bibinfo{person}{Mathieu Reymond}, \bibinfo{person}{Ann Now{\'e}}, {and} \bibinfo{person}{Diederik~M Roijers}.} \bibinfo{year}{2022}\natexlab{}.
\newblock \showarticletitle{Local Advantage Networks for Cooperative Multi-Agent Reinforcement Learning}.
\newblock \bibinfo{journal}{\emph{Proceedings of the 21st International Conference on Autonomous Agents and MultiAgent Systems}} (\bibinfo{year}{2022}).
\newblock


\bibitem[\protect\citeauthoryear{Baker, Kanitscheider, Markov, Wu, Powell, McGrew, and Mordatch}{Baker et~al\mbox{.}}{2020}]%
        {Baker2020Emergent}
\bibfield{author}{\bibinfo{person}{Bowen Baker}, \bibinfo{person}{Ingmar Kanitscheider}, \bibinfo{person}{Todor Markov}, \bibinfo{person}{Yi Wu}, \bibinfo{person}{Glenn Powell}, \bibinfo{person}{Bob McGrew}, {and} \bibinfo{person}{Igor Mordatch}.} \bibinfo{year}{2020}\natexlab{}.
\newblock \showarticletitle{Emergent Tool Use From Multi-Agent Autocurricula}. In \bibinfo{booktitle}{\emph{International Conference on Learning Representations}}.
\newblock


\bibitem[\protect\citeauthoryear{Bengio, Louradour, Collobert, and Weston}{Bengio et~al\mbox{.}}{2009}]%
        {bengio2009curriculum}
\bibfield{author}{\bibinfo{person}{Yoshua Bengio}, \bibinfo{person}{J{\'e}r{\^o}me Louradour}, \bibinfo{person}{Ronan Collobert}, {and} \bibinfo{person}{Jason Weston}.} \bibinfo{year}{2009}\natexlab{}.
\newblock \showarticletitle{Curriculum learning}. In \bibinfo{booktitle}{\emph{Proceedings of the 26th annual international conference on machine learning}}. \bibinfo{pages}{41--48}.
\newblock


\bibitem[\protect\citeauthoryear{Berner, Brockman, Chan, Cheung, D{\k{e}}biak, Dennison, Farhi, Fischer, Hashme, Hesse, et~al\mbox{.}}{Berner et~al\mbox{.}}{2019}]%
        {berner2019dota}
\bibfield{author}{\bibinfo{person}{Christopher Berner}, \bibinfo{person}{Greg Brockman}, \bibinfo{person}{Brooke Chan}, \bibinfo{person}{Vicki Cheung}, \bibinfo{person}{Przemys{\l}aw D{\k{e}}biak}, \bibinfo{person}{Christy Dennison}, \bibinfo{person}{David Farhi}, \bibinfo{person}{Quirin Fischer}, \bibinfo{person}{Shariq Hashme}, \bibinfo{person}{Chris Hesse}, {et~al\mbox{.}}} \bibinfo{year}{2019}\natexlab{}.
\newblock \showarticletitle{Dota 2 with large scale deep reinforcement learning}.
\newblock \bibinfo{journal}{\emph{arXiv preprint arXiv:1912.06680}} (\bibinfo{year}{2019}).
\newblock


\bibitem[\protect\citeauthoryear{Brill and Moore}{Brill and Moore}{2000}]%
        {brill2000improved}
\bibfield{author}{\bibinfo{person}{Eric Brill} {and} \bibinfo{person}{Robert~C Moore}.} \bibinfo{year}{2000}\natexlab{}.
\newblock \showarticletitle{An improved error model for noisy channel spelling correction}. In \bibinfo{booktitle}{\emph{Proceedings of the 38th annual meeting of the association for computational linguistics}}. \bibinfo{pages}{286--293}.
\newblock


\bibitem[\protect\citeauthoryear{Burda, Edwards, Pathak, Storkey, Darrell, and Efros}{Burda et~al\mbox{.}}{2018}]%
        {burda2018large}
\bibfield{author}{\bibinfo{person}{Yuri Burda}, \bibinfo{person}{Harri Edwards}, \bibinfo{person}{Deepak Pathak}, \bibinfo{person}{Amos Storkey}, \bibinfo{person}{Trevor Darrell}, {and} \bibinfo{person}{Alexei~A Efros}.} \bibinfo{year}{2018}\natexlab{}.
\newblock \showarticletitle{Large-scale study of curiosity-driven learning}.
\newblock \bibinfo{journal}{\emph{arXiv preprint arXiv:1808.04355}} (\bibinfo{year}{2018}).
\newblock


\bibitem[\protect\citeauthoryear{Burda, Edwards, Storkey, and Klimov}{Burda et~al\mbox{.}}{2019}]%
        {burda2018exploration}
\bibfield{author}{\bibinfo{person}{Yuri Burda}, \bibinfo{person}{Harrison Edwards}, \bibinfo{person}{Amos Storkey}, {and} \bibinfo{person}{Oleg Klimov}.} \bibinfo{year}{2019}\natexlab{}.
\newblock \showarticletitle{Exploration by random network distillation}. In \bibinfo{booktitle}{\emph{International Conference on Learning Representations}}.
\newblock


\bibitem[\protect\citeauthoryear{Cobbe, Hesse, Hilton, and Schulman}{Cobbe et~al\mbox{.}}{2020}]%
        {cobbe2020leveraging}
\bibfield{author}{\bibinfo{person}{Karl Cobbe}, \bibinfo{person}{Chris Hesse}, \bibinfo{person}{Jacob Hilton}, {and} \bibinfo{person}{John Schulman}.} \bibinfo{year}{2020}\natexlab{}.
\newblock \showarticletitle{Leveraging procedural generation to benchmark reinforcement learning}. In \bibinfo{booktitle}{\emph{International conference on machine learning}}. PMLR, \bibinfo{pages}{2048--2056}.
\newblock


\bibitem[\protect\citeauthoryear{Czarnecki, Jayakumar, Jaderberg, Hasenclever, Teh, Heess, Osindero, and Pascanu}{Czarnecki et~al\mbox{.}}{2018}]%
        {czarnecki2018mix}
\bibfield{author}{\bibinfo{person}{Wojciech Czarnecki}, \bibinfo{person}{Siddhant Jayakumar}, \bibinfo{person}{Max Jaderberg}, \bibinfo{person}{Leonard Hasenclever}, \bibinfo{person}{Yee~Whye Teh}, \bibinfo{person}{Nicolas Heess}, \bibinfo{person}{Simon Osindero}, {and} \bibinfo{person}{Razvan Pascanu}.} \bibinfo{year}{2018}\natexlab{}.
\newblock \showarticletitle{Mix \& match agent curricula for reinforcement learning}. In \bibinfo{booktitle}{\emph{International Conference on Machine Learning}}. PMLR, \bibinfo{pages}{1087--1095}.
\newblock


\bibitem[\protect\citeauthoryear{Fortunato}{Fortunato}{2010}]%
        {fortunato2010community}
\bibfield{author}{\bibinfo{person}{Santo Fortunato}.} \bibinfo{year}{2010}\natexlab{}.
\newblock \showarticletitle{Community detection in graphs}.
\newblock \bibinfo{journal}{\emph{Physics reports}} \bibinfo{volume}{486}, \bibinfo{number}{3-5} (\bibinfo{year}{2010}), \bibinfo{pages}{75--174}.
\newblock


\bibitem[\protect\citeauthoryear{Fudenberg, Nowak, Taylor, and Imhof}{Fudenberg et~al\mbox{.}}{2006}]%
        {fudenberg2006evolutionary}
\bibfield{author}{\bibinfo{person}{Drew Fudenberg}, \bibinfo{person}{Martin~A Nowak}, \bibinfo{person}{Christine Taylor}, {and} \bibinfo{person}{Lorens~A Imhof}.} \bibinfo{year}{2006}\natexlab{}.
\newblock \showarticletitle{Evolutionary game dynamics in finite populations with strong selection and weak mutation}.
\newblock \bibinfo{journal}{\emph{Theoretical population biology}} \bibinfo{volume}{70}, \bibinfo{number}{3} (\bibinfo{year}{2006}), \bibinfo{pages}{352--363}.
\newblock


\bibitem[\protect\citeauthoryear{Girvan and Newman}{Girvan and Newman}{2002}]%
        {girvan2002community}
\bibfield{author}{\bibinfo{person}{Michelle Girvan} {and} \bibinfo{person}{Mark~EJ Newman}.} \bibinfo{year}{2002}\natexlab{}.
\newblock \showarticletitle{Community structure in social and biological networks}.
\newblock \bibinfo{journal}{\emph{Proceedings of the national academy of sciences}} \bibinfo{volume}{99}, \bibinfo{number}{12} (\bibinfo{year}{2002}), \bibinfo{pages}{7821--7826}.
\newblock


\bibitem[\protect\citeauthoryear{Haarnoja, Zhou, Abbeel, and Levine}{Haarnoja et~al\mbox{.}}{2018}]%
        {haarnoja2018soft}
\bibfield{author}{\bibinfo{person}{Tuomas Haarnoja}, \bibinfo{person}{Aurick Zhou}, \bibinfo{person}{Pieter Abbeel}, {and} \bibinfo{person}{Sergey Levine}.} \bibinfo{year}{2018}\natexlab{}.
\newblock \showarticletitle{Soft actor-critic: Off-policy maximum entropy deep reinforcement learning with a stochastic actor}. In \bibinfo{booktitle}{\emph{International conference on machine learning}}. PMLR, \bibinfo{pages}{1861--1870}.
\newblock


\bibitem[\protect\citeauthoryear{Hamilton}{Hamilton}{1964}]%
        {hamilton1964}
\bibfield{author}{\bibinfo{person}{William~D Hamilton}.} \bibinfo{year}{1964}\natexlab{}.
\newblock \showarticletitle{The genetical evolution of social behaviour}.
\newblock \bibinfo{journal}{\emph{Journal of theoretical biology}} \bibinfo{volume}{7}, \bibinfo{number}{1} (\bibinfo{year}{1964}), \bibinfo{pages}{17--52}.
\newblock


\bibitem[\protect\citeauthoryear{Hamilton}{Hamilton}{1972}]%
        {hamilton1972altruism}
\bibfield{author}{\bibinfo{person}{William~D Hamilton}.} \bibinfo{year}{1972}\natexlab{}.
\newblock \showarticletitle{Altruism and related phenomena, mainly in social insects}.
\newblock \bibinfo{journal}{\emph{Annual Review of Ecology and systematics}} \bibinfo{volume}{3}, \bibinfo{number}{1} (\bibinfo{year}{1972}), \bibinfo{pages}{193--232}.
\newblock


\bibitem[\protect\citeauthoryear{Hamming}{Hamming}{1950}]%
        {hamming1950error}
\bibfield{author}{\bibinfo{person}{Richard~W Hamming}.} \bibinfo{year}{1950}\natexlab{}.
\newblock \showarticletitle{Error detecting and error correcting codes}.
\newblock \bibinfo{journal}{\emph{The Bell system technical journal}} \bibinfo{volume}{29}, \bibinfo{number}{2} (\bibinfo{year}{1950}), \bibinfo{pages}{147--160}.
\newblock


\bibitem[\protect\citeauthoryear{Hochreiter and Schmidhuber}{Hochreiter and Schmidhuber}{1997}]%
        {hochreiter1997long}
\bibfield{author}{\bibinfo{person}{Sepp Hochreiter} {and} \bibinfo{person}{J{\"u}rgen Schmidhuber}.} \bibinfo{year}{1997}\natexlab{}.
\newblock \showarticletitle{Long short-term memory}.
\newblock \bibinfo{journal}{\emph{Neural computation}} \bibinfo{volume}{9}, \bibinfo{number}{8} (\bibinfo{year}{1997}), \bibinfo{pages}{1735--1780}.
\newblock


\bibitem[\protect\citeauthoryear{Jaderberg, Czarnecki, Dunning, Marris, Lever, Castaneda, Beattie, Rabinowitz, Morcos, Ruderman, et~al\mbox{.}}{Jaderberg et~al\mbox{.}}{2019}]%
        {jaderberg2019human}
\bibfield{author}{\bibinfo{person}{Max Jaderberg}, \bibinfo{person}{Wojciech~M Czarnecki}, \bibinfo{person}{Iain Dunning}, \bibinfo{person}{Luke Marris}, \bibinfo{person}{Guy Lever}, \bibinfo{person}{Antonio~Garcia Castaneda}, \bibinfo{person}{Charles Beattie}, \bibinfo{person}{Neil~C Rabinowitz}, \bibinfo{person}{Ari~S Morcos}, \bibinfo{person}{Avraham Ruderman}, {et~al\mbox{.}}} \bibinfo{year}{2019}\natexlab{}.
\newblock \showarticletitle{Human-level performance in 3D multiplayer games with population-based reinforcement learning}.
\newblock \bibinfo{journal}{\emph{Science}} \bibinfo{volume}{364}, \bibinfo{number}{6443} (\bibinfo{year}{2019}), \bibinfo{pages}{859--865}.
\newblock


\bibitem[\protect\citeauthoryear{Krakovna, Uesato, Mikulik, Rahtz, Everitt, Kumar, Kenton, Leike, and Legg}{Krakovna et~al\mbox{.}}{2020}]%
        {krakovna2020specification}
\bibfield{author}{\bibinfo{person}{Victoria Krakovna}, \bibinfo{person}{Jonathan Uesato}, \bibinfo{person}{Vladimir Mikulik}, \bibinfo{person}{Matthew Rahtz}, \bibinfo{person}{Tom Everitt}, \bibinfo{person}{Ramana Kumar}, \bibinfo{person}{Zac Kenton}, \bibinfo{person}{Jan Leike}, {and} \bibinfo{person}{Shane Legg}.} \bibinfo{year}{2020}\natexlab{}.
\newblock \showarticletitle{Specification gaming: the flip side of AI ingenuity}.
\newblock \bibinfo{journal}{\emph{DeepMind Blog}} (\bibinfo{year}{2020}).
\newblock


\bibitem[\protect\citeauthoryear{Kuhn}{Kuhn}{2008}]%
        {kuhn2008prisoner}
\bibfield{author}{\bibinfo{person}{Steven Kuhn}.} \bibinfo{year}{2008}\natexlab{}.
\newblock \showarticletitle{Prisoner's dilemma}.
\newblock \bibinfo{journal}{\emph{Stanford Encyclopedia of Philosophy}} (\bibinfo{year}{2008}).
\newblock


\bibitem[\protect\citeauthoryear{Leibo, Hughes, Lanctot, and Graepel}{Leibo et~al\mbox{.}}{2019}]%
        {leibo2019autocurricula}
\bibfield{author}{\bibinfo{person}{Joel~Z Leibo}, \bibinfo{person}{Edward Hughes}, \bibinfo{person}{Marc Lanctot}, {and} \bibinfo{person}{Thore Graepel}.} \bibinfo{year}{2019}\natexlab{}.
\newblock \showarticletitle{Autocurricula and the emergence of innovation from social interaction: A manifesto for multi-agent intelligence research}.
\newblock \bibinfo{journal}{\emph{arXiv preprint arXiv:1903.00742}} (\bibinfo{year}{2019}).
\newblock


\bibitem[\protect\citeauthoryear{Lerner et~al\mbox{.}}{Lerner et~al\mbox{.}}{1958}]%
        {lerner1958genetic}
\bibfield{author}{\bibinfo{person}{Isadore~Michael Lerner} {et~al\mbox{.}}} \bibinfo{year}{1958}\natexlab{}.
\newblock \showarticletitle{The genetic basis of selection.}
\newblock \bibinfo{journal}{\emph{The genetic basis of selection.}} (\bibinfo{year}{1958}).
\newblock


\bibitem[\protect\citeauthoryear{Littman}{Littman}{1994}]%
        {littman1994markov}
\bibfield{author}{\bibinfo{person}{Michael~L Littman}.} \bibinfo{year}{1994}\natexlab{}.
\newblock \showarticletitle{Markov games as a framework for multi-agent reinforcement learning}.
\newblock In \bibinfo{booktitle}{\emph{Machine learning proceedings 1994}}. \bibinfo{publisher}{Elsevier}, \bibinfo{pages}{157--163}.
\newblock


\bibitem[\protect\citeauthoryear{Mills and Beatty}{Mills and Beatty}{1979}]%
        {mills1979propensity}
\bibfield{author}{\bibinfo{person}{Susan~K Mills} {and} \bibinfo{person}{John~H Beatty}.} \bibinfo{year}{1979}\natexlab{}.
\newblock \showarticletitle{The propensity interpretation of fitness}.
\newblock \bibinfo{journal}{\emph{Philosophy of Science}} \bibinfo{volume}{46}, \bibinfo{number}{2} (\bibinfo{year}{1979}), \bibinfo{pages}{263--286}.
\newblock


\bibitem[\protect\citeauthoryear{Mirowski, Pascanu, Viola, Soyer, Ballard, Banino, Denil, Goroshin, Sifre, Kavukcuoglu, et~al\mbox{.}}{Mirowski et~al\mbox{.}}{2017}]%
        {mirowski2016learning}
\bibfield{author}{\bibinfo{person}{Piotr Mirowski}, \bibinfo{person}{Razvan Pascanu}, \bibinfo{person}{Fabio Viola}, \bibinfo{person}{Hubert Soyer}, \bibinfo{person}{Andrew~J Ballard}, \bibinfo{person}{Andrea Banino}, \bibinfo{person}{Misha Denil}, \bibinfo{person}{Ross Goroshin}, \bibinfo{person}{Laurent Sifre}, \bibinfo{person}{Koray Kavukcuoglu}, {et~al\mbox{.}}} \bibinfo{year}{2017}\natexlab{}.
\newblock \showarticletitle{Learning to navigate in complex environments}.
\newblock \bibinfo{journal}{\emph{International Conference on Learning Representations}} (\bibinfo{year}{2017}).
\newblock


\bibitem[\protect\citeauthoryear{Mnih, Kavukcuoglu, Silver, Rusu, Veness, Bellemare, Graves, Riedmiller, Fidjeland, Ostrovski, et~al\mbox{.}}{Mnih et~al\mbox{.}}{2015}]%
        {mnih2015human}
\bibfield{author}{\bibinfo{person}{Volodymyr Mnih}, \bibinfo{person}{Koray Kavukcuoglu}, \bibinfo{person}{David Silver}, \bibinfo{person}{Andrei~A Rusu}, \bibinfo{person}{Joel Veness}, \bibinfo{person}{Marc~G Bellemare}, \bibinfo{person}{Alex Graves}, \bibinfo{person}{Martin Riedmiller}, \bibinfo{person}{Andreas~K Fidjeland}, \bibinfo{person}{Georg Ostrovski}, {et~al\mbox{.}}} \bibinfo{year}{2015}\natexlab{}.
\newblock \showarticletitle{Human-level control through deep reinforcement learning}.
\newblock \bibinfo{journal}{\emph{nature}} \bibinfo{volume}{518}, \bibinfo{number}{7540} (\bibinfo{year}{2015}), \bibinfo{pages}{529--533}.
\newblock


\bibitem[\protect\citeauthoryear{Narvekar, Peng, Leonetti, Sinapov, Taylor, and Stone}{Narvekar et~al\mbox{.}}{2020}]%
        {narvekar2020curriculum}
\bibfield{author}{\bibinfo{person}{Sanmit Narvekar}, \bibinfo{person}{Bei Peng}, \bibinfo{person}{Matteo Leonetti}, \bibinfo{person}{Jivko Sinapov}, \bibinfo{person}{Matthew~E Taylor}, {and} \bibinfo{person}{Peter Stone}.} \bibinfo{year}{2020}\natexlab{}.
\newblock \showarticletitle{Curriculum learning for reinforcement learning domains: A framework and survey}.
\newblock \bibinfo{journal}{\emph{Journal of Machine Learning Research}} \bibinfo{volume}{21}, \bibinfo{number}{181} (\bibinfo{year}{2020}), \bibinfo{pages}{1--50}.
\newblock


\bibitem[\protect\citeauthoryear{Nowak and May}{Nowak and May}{1992}]%
        {nowak1992evolutionary}
\bibfield{author}{\bibinfo{person}{Martin~A Nowak} {and} \bibinfo{person}{Robert~M May}.} \bibinfo{year}{1992}\natexlab{}.
\newblock \showarticletitle{Evolutionary games and spatial chaos}.
\newblock \bibinfo{journal}{\emph{Nature}} \bibinfo{volume}{359}, \bibinfo{number}{6398} (\bibinfo{year}{1992}), \bibinfo{pages}{826--829}.
\newblock


\bibitem[\protect\citeauthoryear{Ohtsuki}{Ohtsuki}{2010}]%
        {ohtsuki2010evolutionary}
\bibfield{author}{\bibinfo{person}{Hisashi Ohtsuki}.} \bibinfo{year}{2010}\natexlab{}.
\newblock \showarticletitle{Evolutionary games in Wright's island model: kin selection meets evolutionary game theory}.
\newblock \bibinfo{journal}{\emph{Evolution: International Journal of Organic Evolution}} \bibinfo{volume}{64}, \bibinfo{number}{12} (\bibinfo{year}{2010}), \bibinfo{pages}{3344--3353}.
\newblock


\bibitem[\protect\citeauthoryear{Ohtsuki, Hauert, Lieberman, and Nowak}{Ohtsuki et~al\mbox{.}}{2006}]%
        {ohtsuki2006simple}
\bibfield{author}{\bibinfo{person}{Hisashi Ohtsuki}, \bibinfo{person}{Christoph Hauert}, \bibinfo{person}{Erez Lieberman}, {and} \bibinfo{person}{Martin~A Nowak}.} \bibinfo{year}{2006}\natexlab{}.
\newblock \showarticletitle{A simple rule for the evolution of cooperation on graphs and social networks}.
\newblock \bibinfo{journal}{\emph{Nature}} \bibinfo{volume}{441}, \bibinfo{number}{7092} (\bibinfo{year}{2006}), \bibinfo{pages}{502--505}.
\newblock


\bibitem[\protect\citeauthoryear{Pathak, Agrawal, Efros, and Darrell}{Pathak et~al\mbox{.}}{2017}]%
        {pathak2017curiosity}
\bibfield{author}{\bibinfo{person}{Deepak Pathak}, \bibinfo{person}{Pulkit Agrawal}, \bibinfo{person}{Alexei~A Efros}, {and} \bibinfo{person}{Trevor Darrell}.} \bibinfo{year}{2017}\natexlab{}.
\newblock \showarticletitle{Curiosity-driven exploration by self-supervised prediction}. In \bibinfo{booktitle}{\emph{International conference on machine learning}}. PMLR, \bibinfo{pages}{2778--2787}.
\newblock


\bibitem[\protect\citeauthoryear{Rogers}{Rogers}{2021}]%
        {rogers2021inclusive}
\bibfield{author}{\bibinfo{person}{Kara Rogers}.} \bibinfo{year}{2021}\natexlab{}.
\newblock \showarticletitle{Inclusive fitness}. In \bibinfo{booktitle}{\emph{Encyclopedia Britannica}}.
\newblock
\urldef\tempurl%
\url{https://www.britannica.com/science/inclusive-fitness}
\showURL{%
\tempurl}


\bibitem[\protect\citeauthoryear{Santos and Pacheco}{Santos and Pacheco}{2005}]%
        {santos2005scale}
\bibfield{author}{\bibinfo{person}{Francisco~C Santos} {and} \bibinfo{person}{Jorge~M Pacheco}.} \bibinfo{year}{2005}\natexlab{}.
\newblock \showarticletitle{Scale-free networks provide a unifying framework for the emergence of cooperation}.
\newblock \bibinfo{journal}{\emph{Physical review letters}} \bibinfo{volume}{95}, \bibinfo{number}{9} (\bibinfo{year}{2005}), \bibinfo{pages}{098104}.
\newblock


\bibitem[\protect\citeauthoryear{Schulman, Wolski, Dhariwal, Radford, and Klimov}{Schulman et~al\mbox{.}}{2017}]%
        {schulman2017proximal}
\bibfield{author}{\bibinfo{person}{John Schulman}, \bibinfo{person}{Filip Wolski}, \bibinfo{person}{Prafulla Dhariwal}, \bibinfo{person}{Alec Radford}, {and} \bibinfo{person}{Oleg Klimov}.} \bibinfo{year}{2017}\natexlab{}.
\newblock \showarticletitle{Proximal policy optimization algorithms}.
\newblock \bibinfo{journal}{\emph{arXiv preprint arXiv:1707.06347}} (\bibinfo{year}{2017}).
\newblock


\bibitem[\protect\citeauthoryear{Silver, Hubert, Schrittwieser, Antonoglou, Lai, Guez, Lanctot, Sifre, Kumaran, Graepel, et~al\mbox{.}}{Silver et~al\mbox{.}}{2018}]%
        {silver2018general}
\bibfield{author}{\bibinfo{person}{David Silver}, \bibinfo{person}{Thomas Hubert}, \bibinfo{person}{Julian Schrittwieser}, \bibinfo{person}{Ioannis Antonoglou}, \bibinfo{person}{Matthew Lai}, \bibinfo{person}{Arthur Guez}, \bibinfo{person}{Marc Lanctot}, \bibinfo{person}{Laurent Sifre}, \bibinfo{person}{Dharshan Kumaran}, \bibinfo{person}{Thore Graepel}, {et~al\mbox{.}}} \bibinfo{year}{2018}\natexlab{}.
\newblock \showarticletitle{A general reinforcement learning algorithm that masters chess, shogi, and Go through self-play}.
\newblock \bibinfo{journal}{\emph{Science}} \bibinfo{volume}{362}, \bibinfo{number}{6419} (\bibinfo{year}{2018}), \bibinfo{pages}{1140--1144}.
\newblock


\bibitem[\protect\citeauthoryear{Suarez, Du, Zhu, Mordatch, and Isola}{Suarez et~al\mbox{.}}{2021}]%
        {suarez2021neural}
\bibfield{author}{\bibinfo{person}{Joseph Suarez}, \bibinfo{person}{Yilun Du}, \bibinfo{person}{Clare Zhu}, \bibinfo{person}{Igor Mordatch}, {and} \bibinfo{person}{Phillip Isola}.} \bibinfo{year}{2021}\natexlab{}.
\newblock \showarticletitle{The Neural MMO Platform for Massively Multiagent Research}.
\newblock \bibinfo{journal}{\emph{Advances in Neural Information Processing Systems}}  \bibinfo{volume}{34} (\bibinfo{year}{2021}).
\newblock


\bibitem[\protect\citeauthoryear{Sutton}{Sutton}{1984}]%
        {sutton1984temporal}
\bibfield{author}{\bibinfo{person}{Richard~Stuart Sutton}.} \bibinfo{year}{1984}\natexlab{}.
\newblock \emph{\bibinfo{title}{Temporal credit assignment in reinforcement learning}}.
\newblock \bibinfo{thesistype}{Ph.D. Dissertation}. \bibinfo{school}{University of Massachusetts Amherst}.
\newblock


\bibitem[\protect\citeauthoryear{Sutton and Barto}{Sutton and Barto}{2018}]%
        {sutton2018}
\bibfield{author}{\bibinfo{person}{Richard~S Sutton} {and} \bibinfo{person}{Andrew~G Barto}.} \bibinfo{year}{2018}\natexlab{}.
\newblock \bibinfo{booktitle}{\emph{Reinforcement learning: An introduction}}.
\newblock \bibinfo{publisher}{MIT press}.
\newblock


\bibitem[\protect\citeauthoryear{Szab{\'o} and Fath}{Szab{\'o} and Fath}{2007}]%
        {szabo2007evolutionary}
\bibfield{author}{\bibinfo{person}{Gy{\"o}rgy Szab{\'o}} {and} \bibinfo{person}{Gabor Fath}.} \bibinfo{year}{2007}\natexlab{}.
\newblock \showarticletitle{Evolutionary games on graphs}.
\newblock \bibinfo{journal}{\emph{Physics reports}} \bibinfo{volume}{446}, \bibinfo{number}{4-6} (\bibinfo{year}{2007}), \bibinfo{pages}{97--216}.
\newblock


\bibitem[\protect\citeauthoryear{Taylor and Stone}{Taylor and Stone}{2009}]%
        {taylor2009transfer}
\bibfield{author}{\bibinfo{person}{Matthew~E Taylor} {and} \bibinfo{person}{Peter Stone}.} \bibinfo{year}{2009}\natexlab{}.
\newblock \showarticletitle{Transfer learning for reinforcement learning domains: A survey.}
\newblock \bibinfo{journal}{\emph{Journal of Machine Learning Research}} \bibinfo{volume}{10}, \bibinfo{number}{7} (\bibinfo{year}{2009}).
\newblock


\bibitem[\protect\citeauthoryear{Tesauro et~al\mbox{.}}{Tesauro et~al\mbox{.}}{1995}]%
        {tesauro1995temporal}
\bibfield{author}{\bibinfo{person}{Gerald Tesauro} {et~al\mbox{.}}} \bibinfo{year}{1995}\natexlab{}.
\newblock \showarticletitle{Temporal difference learning and TD-Gammon}.
\newblock \bibinfo{journal}{\emph{Commun. ACM}} \bibinfo{volume}{38}, \bibinfo{number}{3} (\bibinfo{year}{1995}), \bibinfo{pages}{58--68}.
\newblock


\bibitem[\protect\citeauthoryear{Vinyals, Babuschkin, Czarnecki, Mathieu, Dudzik, Chung, Choi, Powell, Ewalds, Georgiev, et~al\mbox{.}}{Vinyals et~al\mbox{.}}{2019}]%
        {vinyals2019grandmaster}
\bibfield{author}{\bibinfo{person}{Oriol Vinyals}, \bibinfo{person}{Igor Babuschkin}, \bibinfo{person}{Wojciech~M Czarnecki}, \bibinfo{person}{Micha{\"e}l Mathieu}, \bibinfo{person}{Andrew Dudzik}, \bibinfo{person}{Junyoung Chung}, \bibinfo{person}{David~H Choi}, \bibinfo{person}{Richard Powell}, \bibinfo{person}{Timo Ewalds}, \bibinfo{person}{Petko Georgiev}, {et~al\mbox{.}}} \bibinfo{year}{2019}\natexlab{}.
\newblock \showarticletitle{Grandmaster level in StarCraft II using multi-agent reinforcement learning}.
\newblock \bibinfo{journal}{\emph{Nature}} \bibinfo{volume}{575}, \bibinfo{number}{7782} (\bibinfo{year}{2019}), \bibinfo{pages}{350--354}.
\newblock


\bibitem[\protect\citeauthoryear{Waddington}{Waddington}{1968}]%
        {waddington1968towards}
\bibfield{author}{\bibinfo{person}{Conrad~Hal Waddington}.} \bibinfo{year}{1968}\natexlab{}.
\newblock \showarticletitle{Towards a theoretical biology}.
\newblock \bibinfo{journal}{\emph{Nature}} \bibinfo{volume}{218}, \bibinfo{number}{5141} (\bibinfo{year}{1968}), \bibinfo{pages}{525--527}.
\newblock


\bibitem[\protect\citeauthoryear{Wang, Lehman, Clune, and Stanley}{Wang et~al\mbox{.}}{2019}]%
        {wang2019paired}
\bibfield{author}{\bibinfo{person}{Rui Wang}, \bibinfo{person}{Joel Lehman}, \bibinfo{person}{Jeff Clune}, {and} \bibinfo{person}{Kenneth~O Stanley}.} \bibinfo{year}{2019}\natexlab{}.
\newblock \showarticletitle{Paired open-ended trailblazer (poet): Endlessly generating increasingly complex and diverse learning environments and their solutions}.
\newblock \bibinfo{journal}{\emph{arXiv preprint arXiv:1901.01753}} (\bibinfo{year}{2019}).
\newblock


\bibitem[\protect\citeauthoryear{Watkins and Dayan}{Watkins and Dayan}{1992}]%
        {watkins1992q}
\bibfield{author}{\bibinfo{person}{Christopher~JCH Watkins} {and} \bibinfo{person}{Peter Dayan}.} \bibinfo{year}{1992}\natexlab{}.
\newblock \showarticletitle{Q-learning}.
\newblock \bibinfo{journal}{\emph{Machine learning}} \bibinfo{volume}{8}, \bibinfo{number}{3} (\bibinfo{year}{1992}), \bibinfo{pages}{279--292}.
\newblock


\bibitem[\protect\citeauthoryear{Williams}{Williams}{2018}]%
        {williams2018adaptation}
\bibfield{author}{\bibinfo{person}{George~Christopher Williams}.} \bibinfo{year}{2018}\natexlab{}.
\newblock \bibinfo{booktitle}{\emph{Adaptation and natural selection}}.
\newblock \bibinfo{publisher}{Princeton university press}.
\newblock


\bibitem[\protect\citeauthoryear{Zizzo}{Zizzo}{2002}]%
        {zizzo2002measurement}
\bibfield{author}{\bibinfo{person}{Daniel Zizzo}.} \bibinfo{year}{2002}\natexlab{}.
\newblock \showarticletitle{On the measurement of harmony in normal form games}.
\newblock  (\bibinfo{year}{2002}).
\newblock


\end{thebibliography}

\end{document}